\documentclass[conference]{IEEEtran}
\newlength \mycolumnwidth
\usepackage[mathscr]{euscript}
\usepackage{tabularx}
\usepackage{url}
\usepackage{enumitem}
\usepackage{amssymb}
\usepackage{amsmath}
\usepackage{mathtools,xparse}

\usepackage{graphicx}
\usepackage{graphics}
\usepackage{xcolor}
\usepackage{supertabular}
\usepackage{caption}
\usepackage{subcaption}
\usepackage{amsmath}
\usepackage[ruled, linesnumbered, vlined]{algorithm2e}

\usepackage{algpseudocode}
\usepackage{todonotes}

\usepackage{mathtools}
\usepackage{booktabs}

\let\oldnl\nl% Store \nl in \oldnl
\newcommand{\nonl}{\renewcommand{\nl}{\let\nl\oldnl}}% Remove line number for one line
\allowdisplaybreaks
\begin{document}

\title{OPEB: Open Physical Environment Benchmark for Artificial Intelligence}

\author{\IEEEauthorblockN{Hamid Mirzaei}
\IEEEauthorblockA{Dept. of Computer Science\\
University of California, Irvine\\
mirzaeib@uci.edu}
\and
\IEEEauthorblockN{Mona Fathollahi}
\IEEEauthorblockA{Dept. of Computer Science and Engineering\\
University of South Florida\\
mona2@mail.usf.edu}\and
\IEEEauthorblockN{Tony Givargis}
\IEEEauthorblockA{Dept. of Computer Science\\
University of California, Irvine\\
givargis@uci.edu}
}

\maketitle
\begin{abstract}
Artificial Intelligence methods to solve continuous-control tasks have made significant progress in recent years. However, these algorithms have important limitations and still need significant improvement to be used in industry and real-world applications. This means that this area is still in an active research phase. To involve a large number of research groups, standard benchmarks are needed to evaluate and compare proposed algorithms. In this paper, we propose a physical environment benchmark framework to facilitate collaborative research in this area by enabling different research groups to integrate their designed benchmarks in a unified cloud-based repository and also share their actual implemented benchmarks via the cloud. We demonstrate the proposed framework using an actual implementation of the classical mountain-car example and present the results obtained using a Reinforcement Learning algorithm.

\end{abstract}

\section {Introduction}
Recent advancements in using Artificial Intelligence (AI) to solve continuous-control tasks have shown promise as a replacement for conventional control theory to tackle the challenges in emerging complex Cyber-Physical Systems, such as self-driving control, smart urban transportation and industrial robots. An example of AI approaches is Reinforcement Learning (RL). RL algorithms are mostly model-free, meaning that the explicit modeling of the physical system is not required. Also, RL-based agents can work under uncertainty and adapt to the changing environment or objectives. These unique characteristics of RL make it a good candidate to solve the control problem of complex physical systems. However, the RL solutions for continuous control are in their infancy, since there are limitations when applying them in real-world applications. Some examples are unpredictability of agent actions, lack of formal proofs of closed-loop system stability and not being able to transfer learning from one task to other tasks with slight modifications. This calls for extensive research to address these limitations and design RL and other AI algorithms that can be used in real-world applications. 

While there are a number of widely-used benchmarks in different computing domains, for example MiBench \cite{guthaus2001mibench} for embedded processing and ImageNet \cite{deng2009imagenet} for computer vision, the available AI benchmarks are very limited. This makes conducting research in AI difficult and expensive. Moreover, since there are not many available standard benchmarks, it is hard to evaluate and compare newly proposed AI algorithms. One of the reasons for the lack of AI benchmarks is the interactive nature of dynamical systems. In other words, while it is possible for many other domains to record and label datasets and make them publicly available, AI benchmark developers should provide an interactive ``environment'' which the AI agent must be able to interact with by applying actions and gathering the new system state (or observation) along with reward signals. This makes AI benchmark development a challenging task. Nevertheless, significant progress has been made recently towards building simulation/emulation based AI benchmarks such as OpenAI Gym and OpenAI Universe \cite{brockman2016openai}.

Although the recently developed AI benchmarks enable the researchers to apply their algorithms on a vast variety of different artificial environments, such as PC games or physical systems simulations, real-world physical environments such as industrial robots and self-driving cars are only available to a limited number of groups in big institutes due to the high costs of manufacturing and maintenance of those environments. The lack of physical benchmarks slows down the research progress in developing AI algorithms that can address challenges that usually exist in the real-world such as sensor noise and delay, processing limitations, communicational bandwidth, etc., and can be used in emerging Internet-of-things (IoT) and Cyber-Physical systems.

In this paper, we propose the Open Physical Environment Benchmark (OPEB) framework to integrate different physical environments. Similar to OpenAI Gym, in our approach a unified interface of the environments is proposed that enables research groups to integrate their physical environment designs to OPEB regardless of the details involved in the hardware/software design and implementation. To achieve the main goals of universality and affordability, we propose leveraging 3D printing technology to build the customized mechanical parts required in the environments and using low-cost generic hardware components such as bolts, ball bearings, etc. We also use popular and affordable embedded processing platforms, such as the Raspberry Pi \cite{upton2014raspberry}, which is a promising processing solution for IoT and Industry 4.0.

Furthermore, the users are not only able to replicate physical environments using OPEB, but they can also share the implemented environment on the cloud enabling other users to evaluate their algorithms on the actual physical environment. This feature results in higher availability of physical benchmarks and facilitates collaborative research to design robust AI algorithms that can be applied on different realizations of an environment with slight variations in the physical properties of the hardware components. Since OPEB is based on low-cost fabrication solutions, it can be used for educational purposes for IoT, control, AI and other related courses.

The remainder of this paper is organized as follows. In Section \ref{sec:backgr-relat-work}, we review the background and some related works in AI benchmarks. In Section \ref{sec:open-phys-envir}, the elements of a physical environment are introduced and it is explained how the required artifacts are provided in OPEB to replicate a physical environment. In Section \ref{sec:example-impl-class}, an example implementation of an OPEB, i.e., the classical mountain-car problem, is described, and the results of the experiments that are performed on the physical system using an RL-based method is presented in Section \ref{sec:results}. Finally, conclusions are presented in Section \ref{sec:concl}.

\section {Background and related work}\label{sec:backgr-relat-work}
In this section, we first review existing literature about solving real-world tasks using AI algorithms. Next, we review recent simulation-based AI benchmarks that are widely used in academia. Finally, we review the related research projects to provide real-world benchmarks in robotic applications.

Using RL as a replacement for conventional control theory is an emerging trend in Cyber-Physical systems. In \cite{navarro2012real} an RL algorithm is proposed to autonomously navigate a humanoid Nao robot into a docking station used for recharging. An RL model is proposed in \cite{levine2016learning} to learn hand-eye coordination for grasping objects in an environment of robotic manipulators. In \cite{mattner2012learn}, RL methods have been applied on an actual cart-pole system to balance the pole. Researchers are exploring AI algorithms as a way to simplify and speed up the programming of industrial robots in factories. Fanuc \cite{katsuki2017machine}, the world's largest maker of industrial robots, has used RL methods to train robots to precisely pick up a box and put it in a container. In the automotive industry, authors in \cite{pecka2017controlling} have proposed an RL-based approach to control robot morphology (flippers) to move over rough terrains that exist in Urban Search and Rescue missions. 

Access to these physical environments (hardwares/robots) is not feasible for a lot of research groups. This hinders partnerships and cooperation between academia and industry. In this paper, for the first time, we propose the idea of providing low-cost and easy-to-construct physical environments that allow researchers and students to implement, evaluate and compare their AI algorithms on standardized benchmarks.

In a dynamic AI problem, the state of the environment depends on the actions that are chosen by the agent. This makes it almost impossible to store the environment as a fixed dataset similar to the supervised machine learning paradigm. Therefore, to facilitate reproducible research and accelerate the pace of education, researchers in this community are trying to design a standard programming interface for reinforcement-learning experiments.

One of the earliest efforts to design a standard tool is RL-Glue \cite{tanner2009rl} which has been used for RL courses in several universities and to create experiments for scientific papers. A more recent effort, RLPy \cite{geramifard2015rlpy}, is a software framework written in python that has focused on value-function-based methods with linear function approximation using discrete actions. ALE \cite{bellemare13arcade} is another software framework designed to make it easy to develop agents that play different genres of Atari 2600 games.  

OpenAI Gym \cite{brockman2016openai} is the most recent and comprehensive toolkit for developing AI algorithms. It provides a diverse suite of environments that range from classic control to 2D and 3D robots. It is designed to let the users evaluate the proposed AI algorithms with little background in AI. Researchers can compare the performance of their proposed algorithm with other approaches' scores reported on the scoreboard. These solutions are very effective in advancement of research and education within simulated environments because it is usually expensive and more challenging to implement AI algorithms in real-world scenarios. 

Most similar to our work is \cite{sheh2014open} that has proposed an open hardware design for academic and research robots. They have leveraged 3D printing technology to allow users to create all required components except electronics parts. All basic code and libraries have been released under the GNU General Public License. Authors in \cite{costa2016design} have made their research on aquatic swarm robots reproducible by providing the 3D printing models, CNC milling files and the developed software on Raspberry Pi. In this paper, we propose a framework that can be used to produce an arbitrary number of physical environments, not limited to robots. Contrary to the mentioned works where a specific physical environment is introduced, a unified benchmark framework is proposed in this paper to integrate a variety of physical environments. In other words, research groups can contribute by sharing their physical environment blueprints using the proposed framework. The other contribution is that users are able to share their actual implementation via a web-based software on the cloud to be used by others for research and education purposes.

\section {Open Physical Environment Benchmark (OPEB)} \label{sec:open-phys-envir}
In this section, we describe our OPEB framework. First, the elements of a physical environment (PE) are introduced and the requirements for each element are discussed. Next, we will explain how the required components to replicate the PE are encapsulated in OPEB and also how the actual implementation can be shared to other users on the cloud.

\vspace{-5pt}
\subsection {Physical Environment Elements} 
The PE consists of the following elements:
\begin{itemize}
\item Mechanical parts and structures
\item Electromechanical components
\item Electrical components
\item Embedded processing unit
\item Embedded software
\end{itemize}

To achieve the goal of affordability and universality of PE implementation, the physical parts should include either generic mechanical hardware such as bolts, ball-bearings, etc., or the parts that can be easily printed using a 3D printer. The electromechanical parts such as actuators, dc motors or transducers should be generic parts that can be easily found all over the world. For example, low cost hobby electromechanical parts can be used to build a PE. To drive and interface the electromechanical parts, some electrical parts such as motor drives should be included in the PE. Additionally, to measure the physical quantities, some sensors are required. Examples of such sensors are digital camera, thermometer and proximity sensor.

The embedded processing unit is needed to perform basic required tasks to run the environment such as timing, reading the sensors' outputs and the required signal processing, producing the environment observation, applying the action calculated by the AI algorithm, sending the monitoring data over the network to the monitoring node locally or over the cloud and running the AI algorithm. These tasks are implemented by the embedded software developed for the PE. All of the software components are provided by the OPEB except the AI algorithm which is developed by the PE user.

Emerging single-board embedded computing platforms can be used as the embedded processing unit in PE. Some examples of these solutions are Raspberry Pi \cite{upton2014raspberry}, C.H.I.P. computer \cite{chip} and Arduino \cite{banzi2014getting} platforms. Using a dedicated embedded processor instead of a general purpose computer reduces the cost of deployment of multiple instances of the PE on the cloud and simplifies interfacing the electrical and electromechanical elements because most of these platforms have on-board I/O capabilities.

\vspace{-5pt}
\subsection{OPEB Components}

\begin{figure}[!t]
  \centering
  \includegraphics[width= 0.9 \mycolumnwidth]{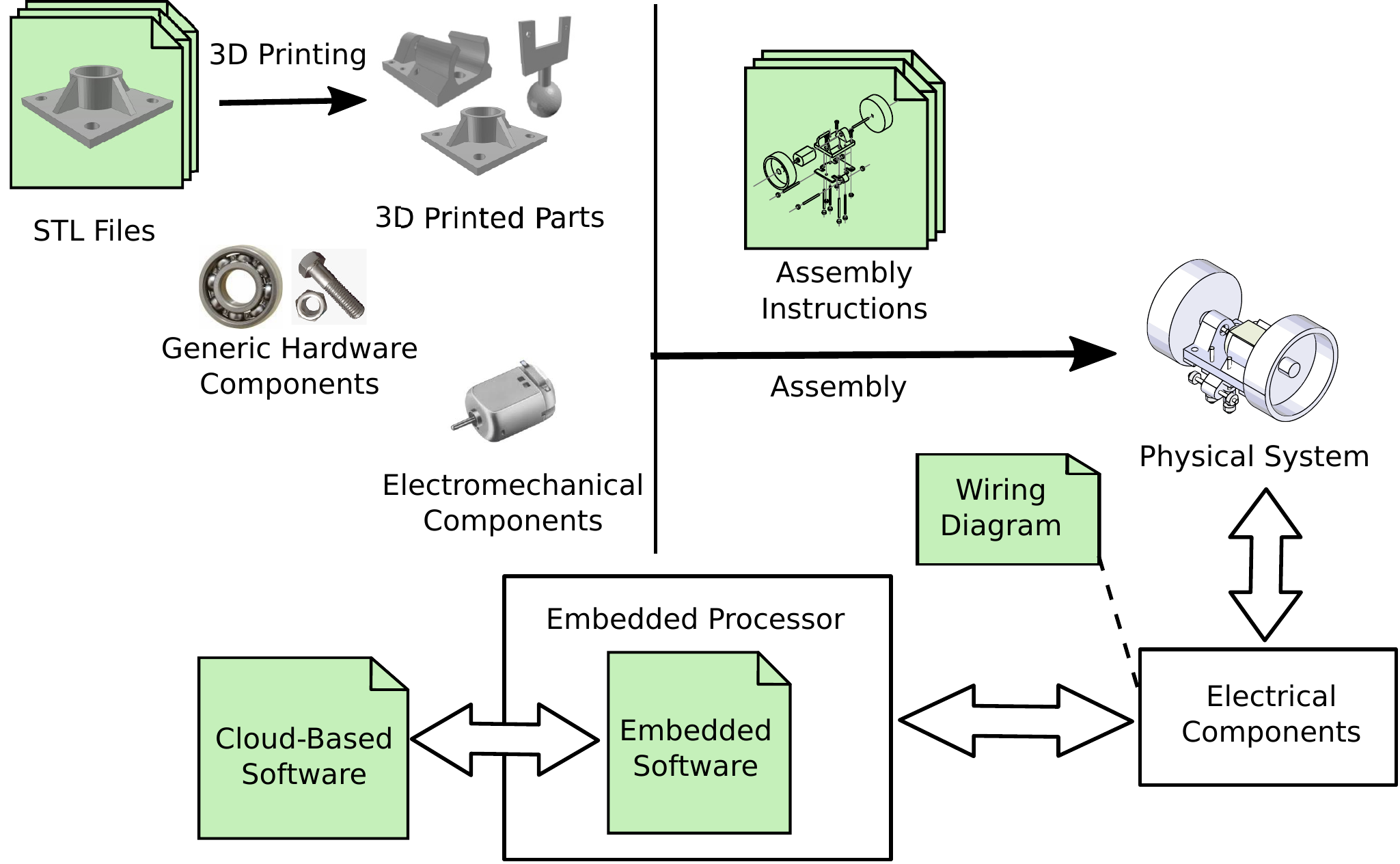}
  \caption{OPEB framework components for each environment. Green blocks are provided in OPEB. All other components listed and specified in OPEB.}
  \label{fig:opeb}
\end{figure}

In Fig. \ref{fig:opeb}, the different components of OPEB for each environment are shown. To realize an environment consisting of the elements listed in the previous subsection, the following components are provided in OPEB for that specific environment:
\begin{itemize}
\item Parts that should be 3D printed in STL \cite{grimm2004user} format.
\item List of materials of the generic mechanical hardware.
\item Diagrams and instructions required for mechanical structure assembly.
\item List of electrical and electromechanical components.
\item List of embedded processing units and peripherals.
\item Wiring diagram of the electrical components.
\item PE control and monitoring Embedded software.
\item Web application for the cloud-based sharing of the PE.
\end{itemize}

The customized mechanical parts required by a PE are included in OPEB as 3D models in STL format that can be easily fabricated using a 3D printer. The specifications of other parts that are not printable or can be selected from off-the-shelf products are provided in OPEB. However, these parts are generic mechanical hardwares that are supplied by many manufacturers around the world.

Besides the information provided to obtain or fabricate the components, OPEB includes the complete instructions and diagrams to assemble the mechanical structures of the PE. The main goal of OPEB is that the environments can be reproduced with minimum discrepancy across different implementations. To achieve this goal, the user should be able to build the whole environment using the provided components in the OPEB without ambiguity. On the other hand, the instruction assembly should be of low complexity and easy to follow to be usable by users with different levels of expertise. For this purpose, a step-by-step assembly instruction approach proposed in \cite{agrawala2003designing} is employed for the mechanical and electromechanical parts.

Electrical and electromechanical parts, including actuators, sensors, processing units and drivers are usually selected from off-the-shelf products. The list of needed components and their specification are listed in OPEB for each environment. Also, unambiguous wiring diagrams are provided for electrical interconnections.

After building the hardware components, the embedded software should be deployed on the embedded processing unit. The embedded software is included in OPEB and can be deployed using installation manuals. To enable the OPEB users to evaluate their algorithms using different PEs, a standard API is defined similar to OpenAI Gym environments. More specifically, the AI agent can interact with the PE using functions that apply actions and returns the environment observations and reward signal. Furthermore, the environment can be reset to the initial state using the PE API.

Finally, the back-end and front-end software components are provided that enable the OPEB users to deploy their implemented PE over the cloud. Using this web-based application, other users can use the PE to upload and run their AI algorithms on the physical system and see the evaluation reports such as accumulated score over time and record the videos of the PE that runs their algorithm.

\section{Example implementation: Classical Mountain-Car Example} \label{sec:example-impl-class}
In this section, we discuss the process of developing an example OPEB environment, i.e., the Mountain-Car example, to demonstrate the methods mentioned in Section \ref{sec:open-phys-envir}. 

\begin{figure}[!t]
  \centering
  \includegraphics[width= 0.6 \mycolumnwidth]{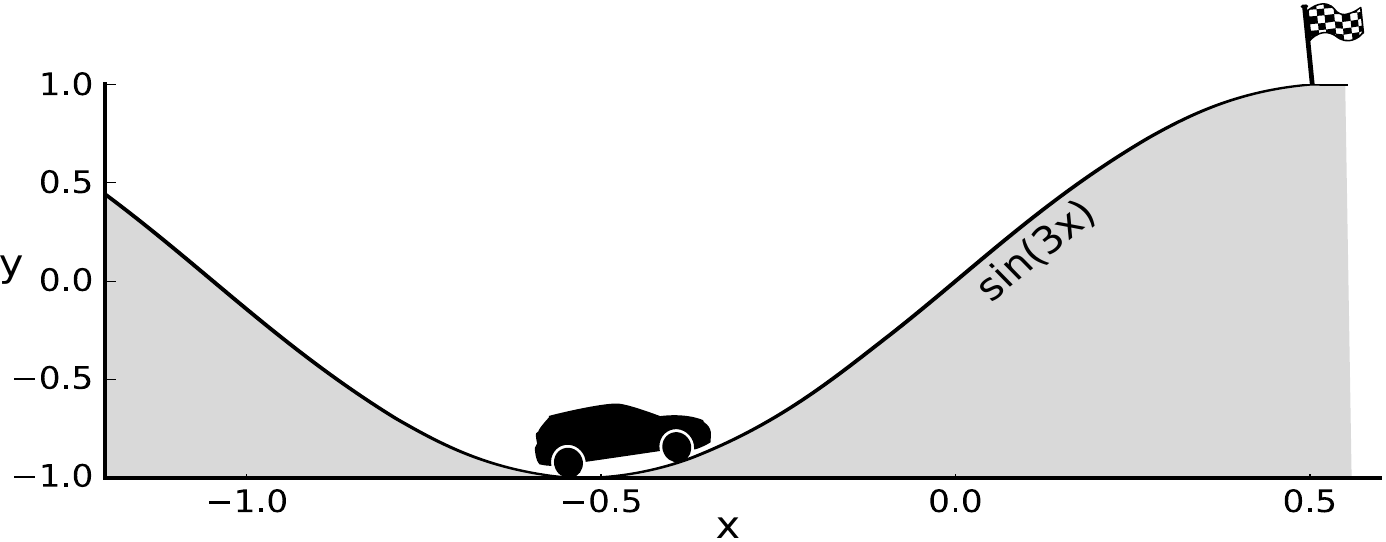}
  \caption{Mountain Car example}
  \label{fig:mountcar}
\end{figure}

In the Mountain Car example, which is first introduced in \cite{moore1990efficient}, the goal is to control the acceleration of a car inside a valley in order to move it to the top of the mountain (Fig. \ref{fig:mountcar}). However, the maximum acceleration of the car is limited and it can not be driven to the top of mountain in a single pass and the car has to go back and forth a number of times to get enough momentum to reach to the desired destination. An AI solution based on Q-learning and tile coding approximation is presented in \cite{sutton1998reinforcement} for this example with a fast convergence in a couple of hundred episodes. However, several simplifying assumptions are made in the original mountain car example including simplified dynamics equations, exact measurements without noise and nonlineariy, no sensor or processing delays and car motion with no friction and no slipping. The last assumption makes the learning process a fairly easy task since the kinetic energy delivered by the car's motor is preserved in the system. Consequently, the car can endlessly swing in the valley and the AI agent can make gradual progress towards the goal by increasing the swing range bit-by-bit using successive actions. In a real-world situation, none of these assumptions hold and the agent has to learn a successful policy in a limited time since the car is going to stop after a few swings. The mentioned limitations justify the importance of physical benchmarks that can evaluate the AI algorithms which are useful in real-world applications, for example industrial robotics or self-driving vehicles.

\vspace{-5pt}
\subsection{Mechanical Structures}
\vspace{-1pt}

\begin{figure}[!t]
  \centering
  \includegraphics[width= 0.26 \mycolumnwidth]{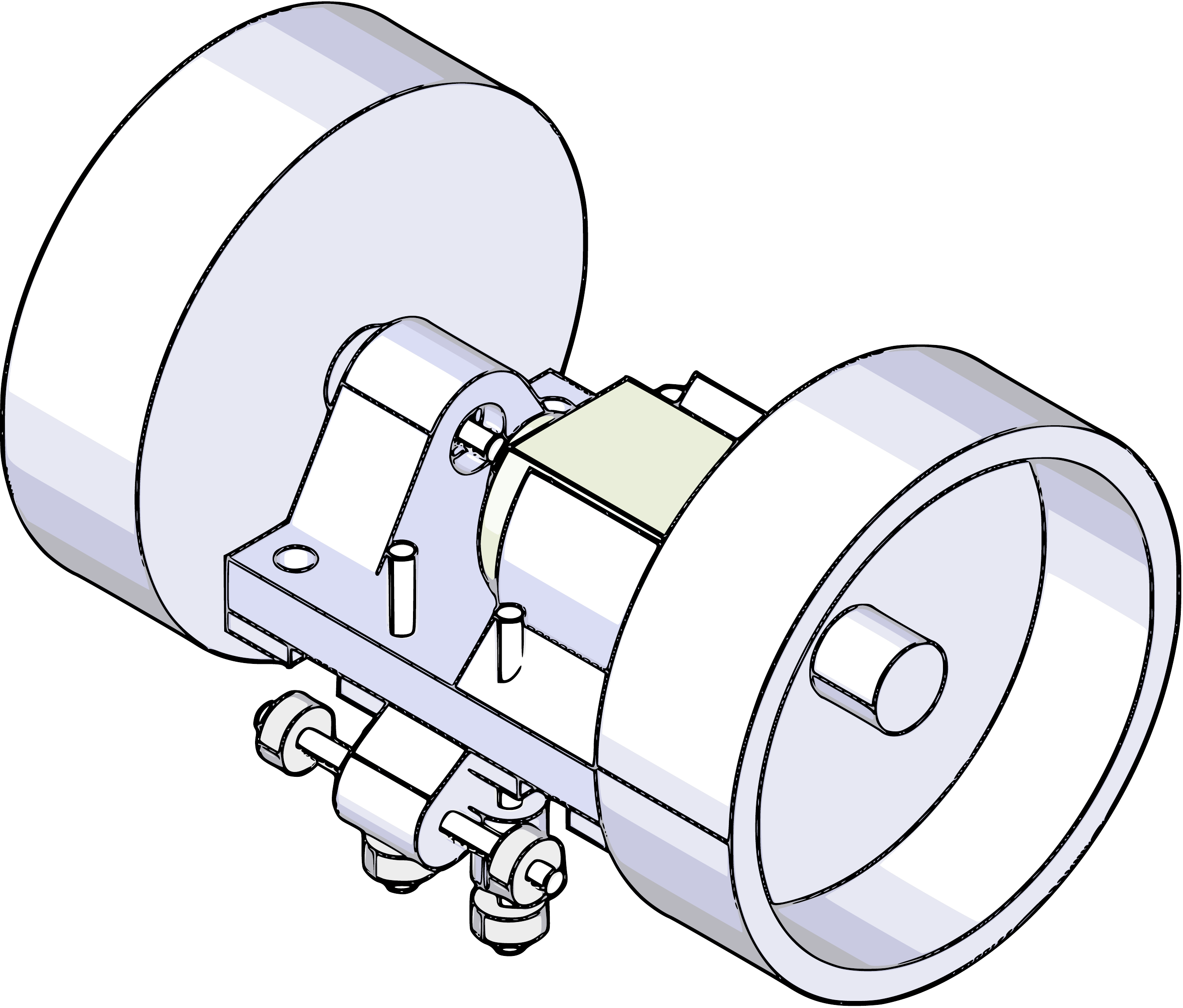}
  \caption{Car Assembly in the MC-OPEB.}
  \label{fig:car}
\end{figure}

The MC-OPEB consists of two mechanical structures: Car and Mountain rail. The car, which is shown Fig. \ref{fig:car}, consists of only two large wheels because a car with two pairs of rear and front wheels might entangle around the positions of the path that have low radius of curvature. Also, using only two wheels results in less overall car weight which enables us to use a low power motor and simplifies the design or selection of electrical parts such as motor drive and power supply. Moreover, to prevent the motor from spinning and to constrain the car to move inside the mountain rail, 8 pieces of small ball-bearings are embedded in the car structure using short metal bars.

\begin{figure}[!t]
  \centering
  \includegraphics[width= 0.5 \mycolumnwidth]{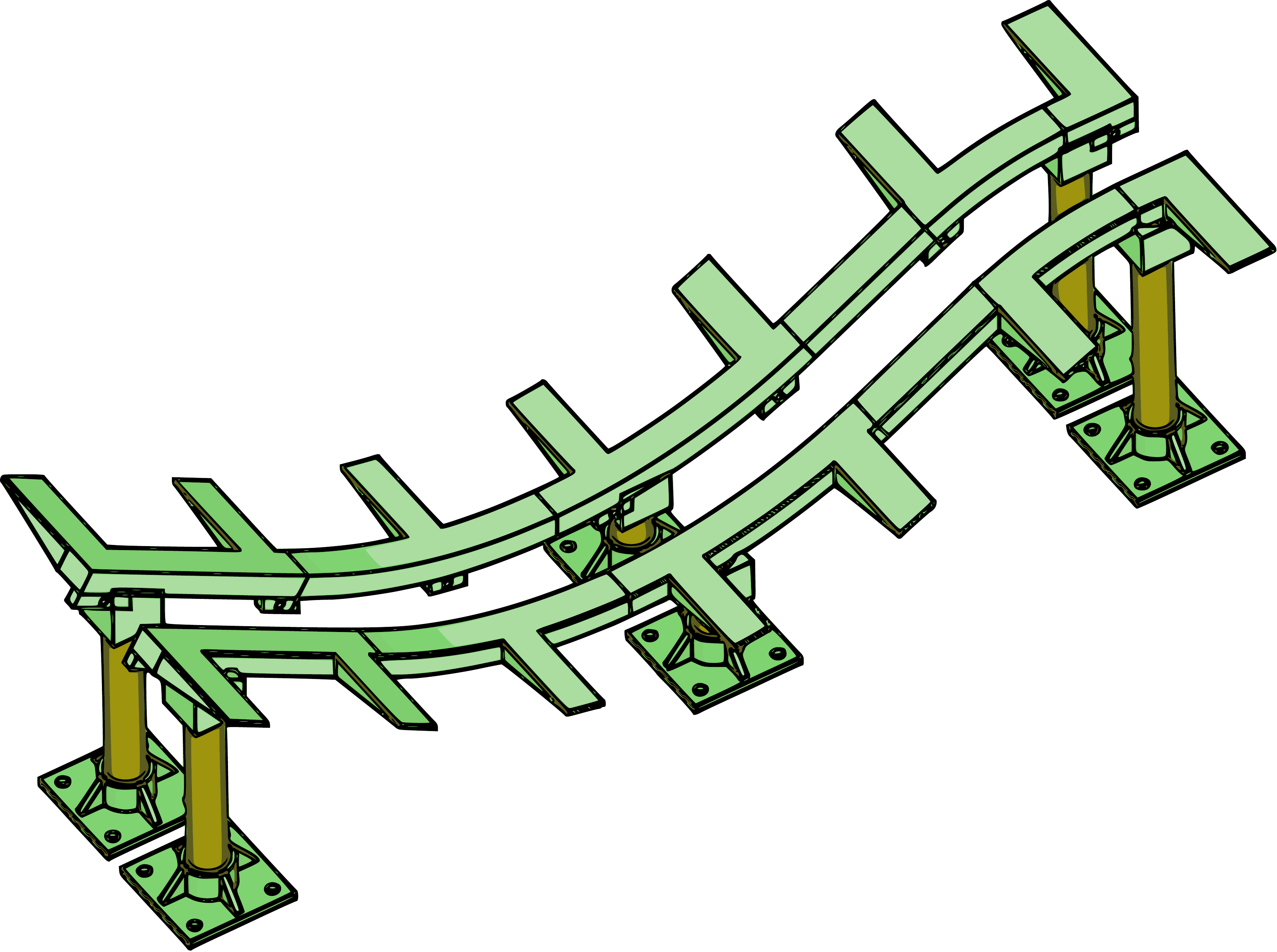}
  \caption{Mountain Rail Assembly in the MC-OPEB.}
  \label{fig:rail}
\end{figure}

\begin{figure}[!b]
  \centering
  \includegraphics[width= .6 \mycolumnwidth]{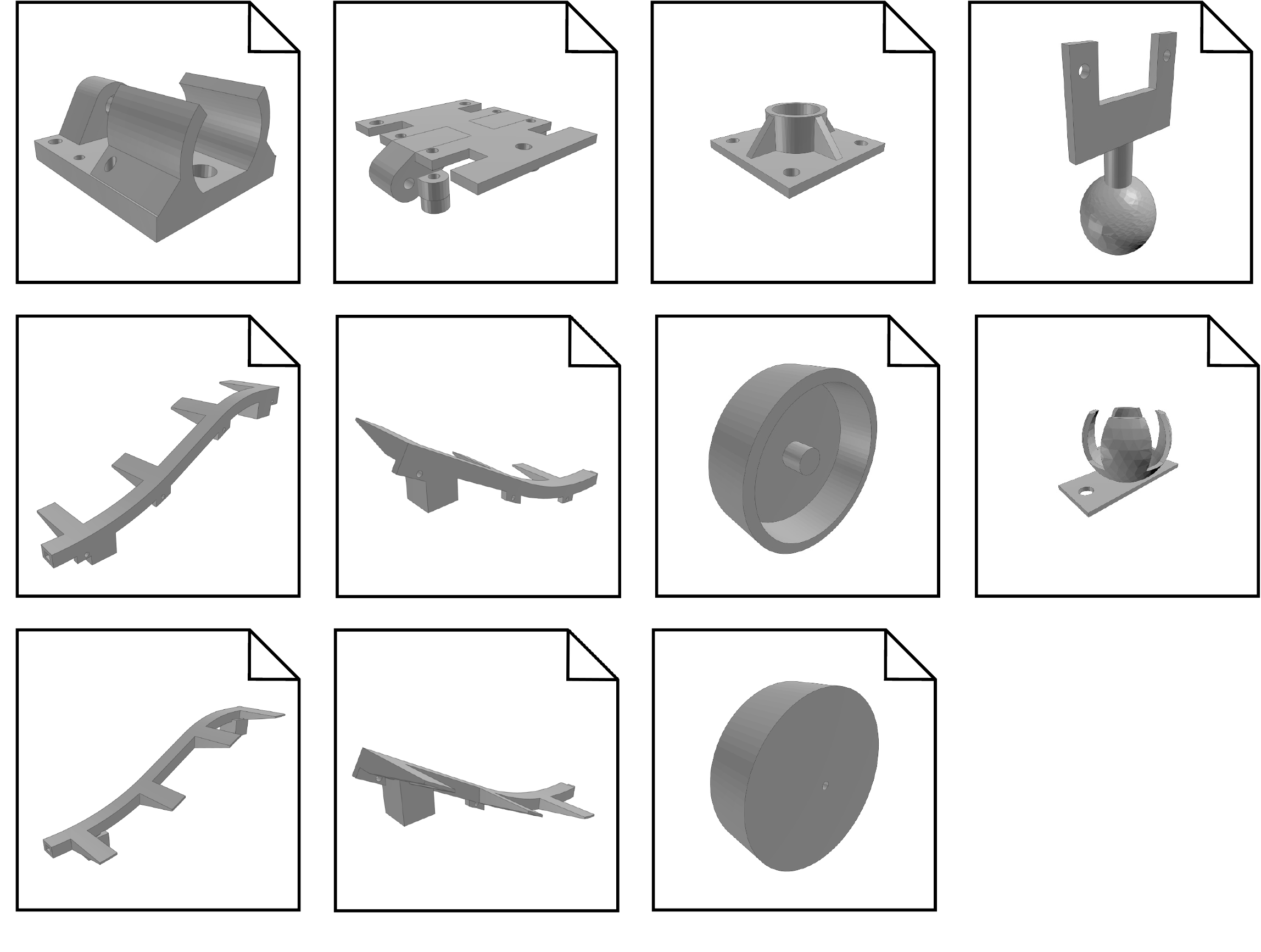}
  \caption{STL files included for MC-OPEB for all required 3D printed parts.}
  \label{fig:stls}
\end{figure}

Each side of the mountain rail, which is shown in Fig. \ref{fig:rail}, is divided to two smaller parts to make them printable using 3D printers with small beds. Additionally, the whole rail surface is not printed to preserve filament. A flexible cardboard should be placed on the support bars attached to the rail structure. The complete STL set of the 3D printed objects are shown in Fig. \ref{fig:stls} and the set of required hardware is listed in Table \ref{tab:bom}.

\begin{table}[!b]
\centering
\caption{List of Materials of required generic hardware parts}
\label{tab:bom}
\begin{tabular}{|l|l|}
\hline
Item                   & Quantity \\ \hline
3mmx10mm bolt and nut  & 29       \\ \hline
32mmx2mm steel bar     & 7        \\ \hline
2x6x2.5mm ball bearing & 8        \\ \hline
10mmx100mm wooden bar  & 1        \\ \hline
\end{tabular}
\end{table}

\begin{figure}[!b]
  \centering
  \includegraphics[width= .5 \mycolumnwidth]{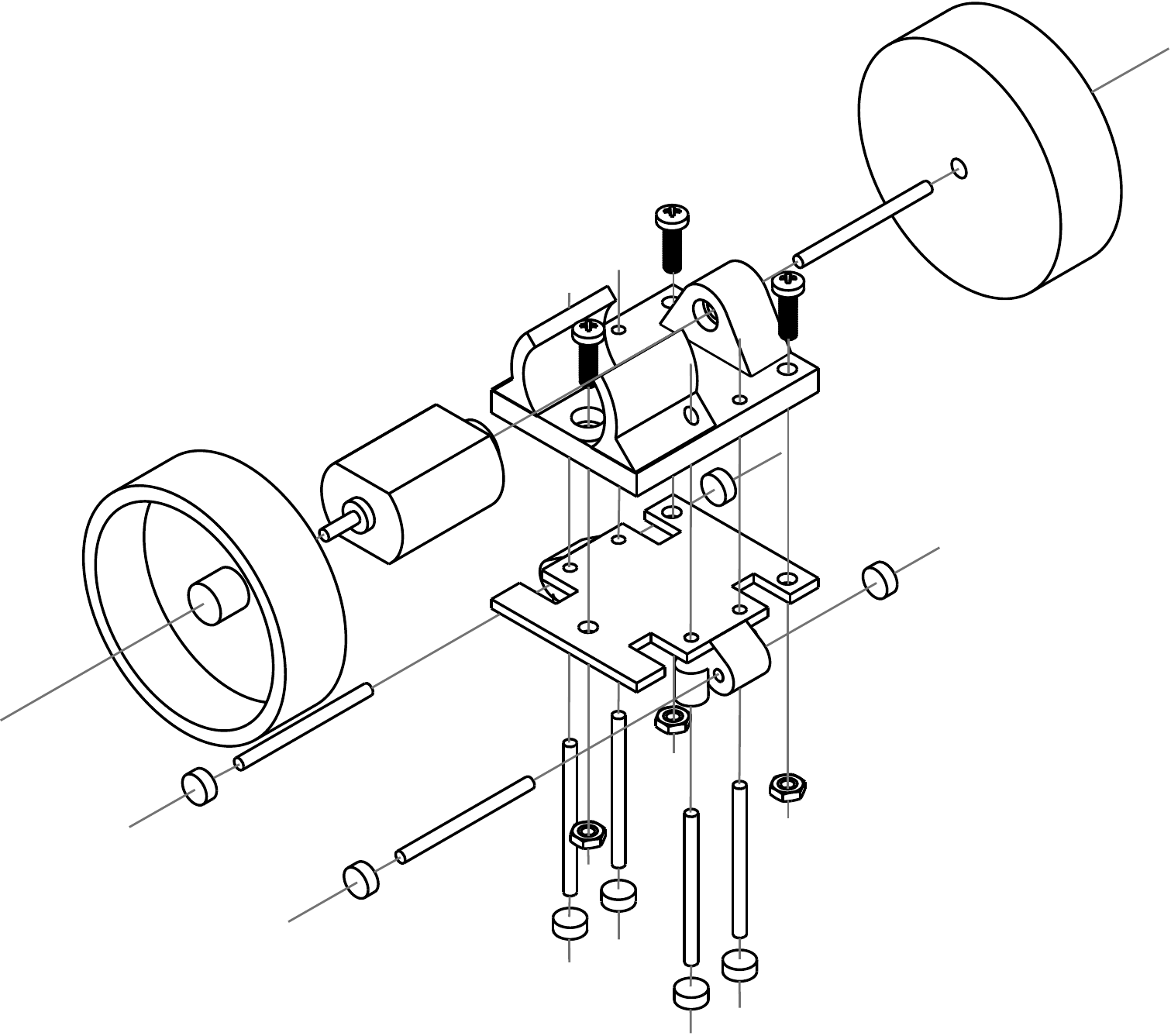}
  \caption{Exploded-view of car assembly as an example of assembly instruction diagrams in MC-OPEB.}
  \label{fig:expl}
\end{figure}

An example of assembly instruction documents is provided in Fig. \ref{fig:expl} which shows the exploded-view diagram of car assembly. The assembly instruction includes the step-by-step action diagrams as explained in \cite{agrawala2003designing}.

\vspace{-5pt}
\subsection{Electromechanical Parts}
\vspace{-1pt}
The only electromechanical part needed for MC-OPEB is the widely-used and low-cost 1.5-3 (V) hobby motor. To reduce the friction and simplify the mechanical design this motor is directly coupled to one of the large wheels on the car. Also, no transducers, such as potentiometer or a shaft encoder is coupled to the motor to reduce the weight of the car and overall cost of MC-OPEB. 

\vspace{-5pt}
\subsection{Electrical Parts}
\vspace{-1pt}
The required electrical parts are: motor driver, two 5V power supplies for the Raspberry Pi board and driving the motor, and Raspberry pi camera. We have used low-cost HG7881 motor drive with PWM inputs. Since the Raspberry Pi has two on-board pwm outputs we can directly connect it to the motor drive without any additional interfacing circuit.

The Raspberry Pi camera is used to measure the motion quantities of the car, i.e., position and speed. The captured image of the car also can be used to evaluate emerging deep reinforcement learning algorithms that can control a physical system only by raw visual data.

\vspace{-5pt}
\subsection{Embedded Processing Unit}
\vspace{-1pt}
We have used ``Raspberry Pi Zero W''  platform which is a powerful and affordable processing unit for different embedded applications. 

\vspace{-5pt}
\subsection{Embedded Software}
\vspace{-1pt}
The Embedded software used in MC-OPEN is a C++ program that is executed on the Raspbian Jessie OS. The embedded software is responsible for implementing the 0.01(s) control timing, capturing and processing the camera image, running the AI routine supplied by the environment user, applying the motor voltage command using PWM outputs, sending monitoring data consisting of instantaneous speed, position and other status variables, running the learned policy and recording the performance video upon user's request.

The camera image is post processed to calculate the position and speed of the car which are the observations of the MC-OPEB. First, the HSV pixel values are filtered by some fixed thresholds to extract the pixels of the yellow marker attached to the car. Next, the spatial moments of filtered pixels are calculated and used to obtain the single $(x,y)$ coordinate of the car. To reduce the noise and estimate the car's speed, a linear Kalman filter is implemented in the embedded software.

\begin{algorithm}[!t]
  \scriptsize
 \KwData{$x, v$ \Comment {Instantaneous car position(x) and speed(v)}}
 \KwResult{$a$ \Comment {action(a): acceleration direction}}
 \If{$|v| < 50$} {
    Choose $a \leftarrow \text{left}$ or $a \leftarrow \text{right}$ randomly with same probability.;
 } \Else {
   \If{$v > 0$} {
     $a \leftarrow \text{left}$;
   } \Else {
     $a \leftarrow \text{right}$;
   }
 }

 \caption{\footnotesize Hand-engineered policy for the mountain-car environment}
 \label{alg:handeng}
\vspace{-5pt}
\end{algorithm}
\vspace{-5pt}

\subsection{Web Application}
The web application is an optional component that can be run on a secondary general purpose computer. Using the web application, the MC-OPEB user can see the monitoring data online and share the implemented physical environment on the cloud. The cloud user can upload a c++ routine that implements any custom AI application and evaluate the algorithm performance using the web application. The cloud user can also pause the learning and run the learned policy and see the recorded view of the actual AI algorithm performance.

\begin{figure}[!t]
  \centering
  \includegraphics[width=0.7\mycolumnwidth]{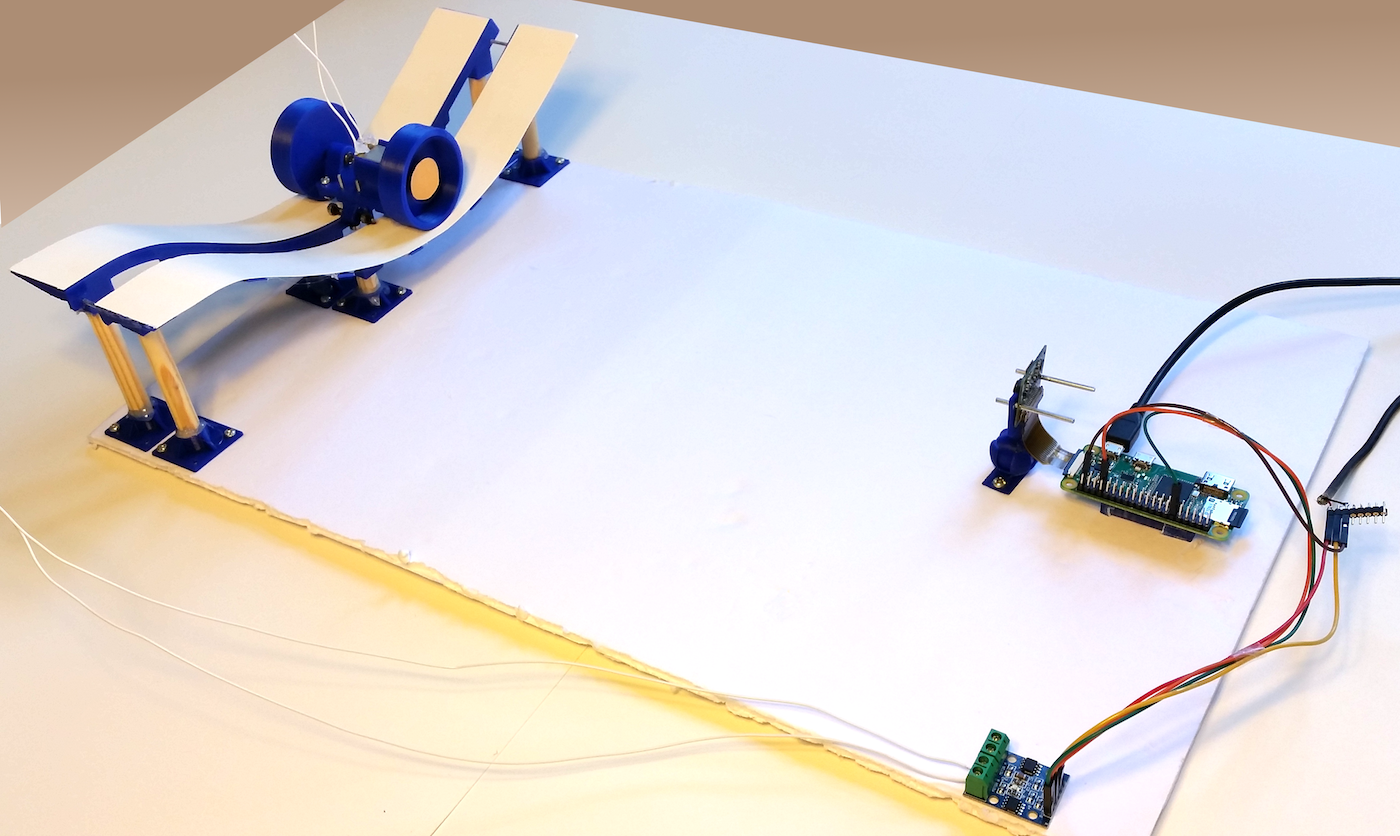}
  \caption{Actual Implementation of MC-OPEB.}
  \label{fig:real}
\end{figure}

Fig. \ref{fig:real} shows a picture of the actual MC-OPEB. In the next section, we show the results of running a reference algorithm and an RL-based algorithm on the built environment.

\section{Results}\label{sec:results}
In this section, we present the results of the experiments performed on the MC-OPEB to show the effectiveness of a low-cost PE to perform real-world experiments using AI methods. The objective is to move the car to a certain height on the left side of the rail which corresponds to 80 pixel displacement of the car to the left in the captured image. The reward is defined is as -1 for all the sampling times that the car has not reached the destination. Each episode starts from the car being at the bottom of the valley and ends when it reaches the desired height on the left side. Therefore, the total reward which is the RL ``return'' is proportional to the negated total episode time. The action is the car's acceleration direction assuming that the car moves with the maximum acceleration and only changes the direction of the acceleration.

\vspace{-5pt}
\subsection{Reference Solution}
\begin{figure}[!t]
  \centering
  \includegraphics[width=0.75 \mycolumnwidth]{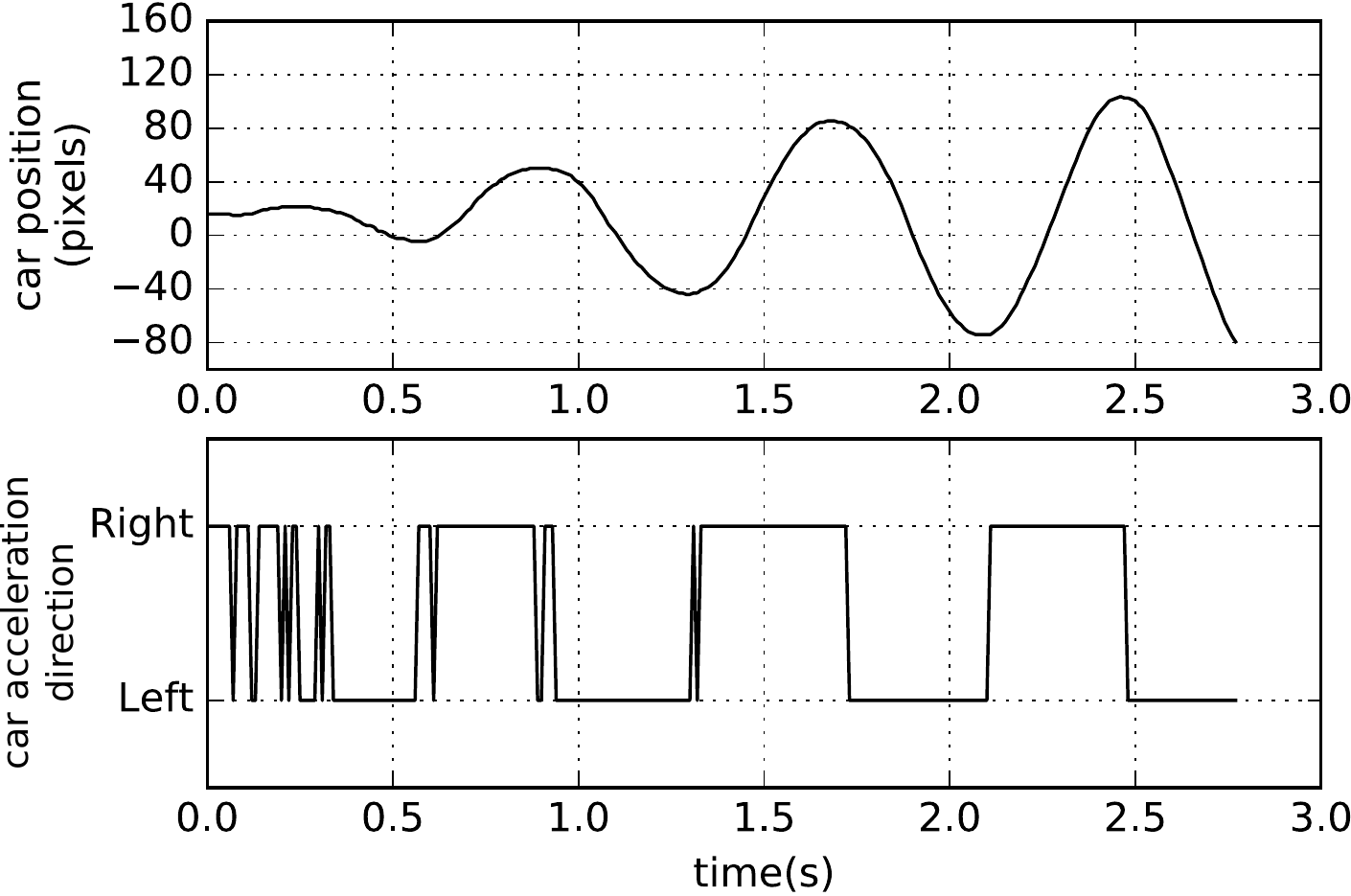}
  \caption{The upper plot represents the car position vs. time. The lower plot represents the car acceleration command computed by the hand-engineered algorithm. After a few swings, the designed algorithm is able to move the car to desired height on the left side which corresponds to -80 pixel coordinate of the car in the captured image.}
  \label{fig:ref}
\end{figure}

\vspace{-1pt}
To ensure the possibility of moving the car from the lowest point in the valley to some certain height by any algorithm, a hand-engineered solution is proposed in Algorithm \ref{alg:handeng}. The performance of the AI-based solution can be compared with the reference solution to evaluate the AI algorithm. Fig. \ref{fig:ref} shows the result of the reference solution.

\vspace{-5pt}
\subsection{AI-based solution}
\vspace{-1pt}
The Q-learning algorithm with tile-coding function approximation is used to show that the proposed MC-OPEB can be used to evaluate AI algorithms on a physical environment in real-time.

\begin{figure}[!t]
  \centering
  \includegraphics[width=0.75 \mycolumnwidth]{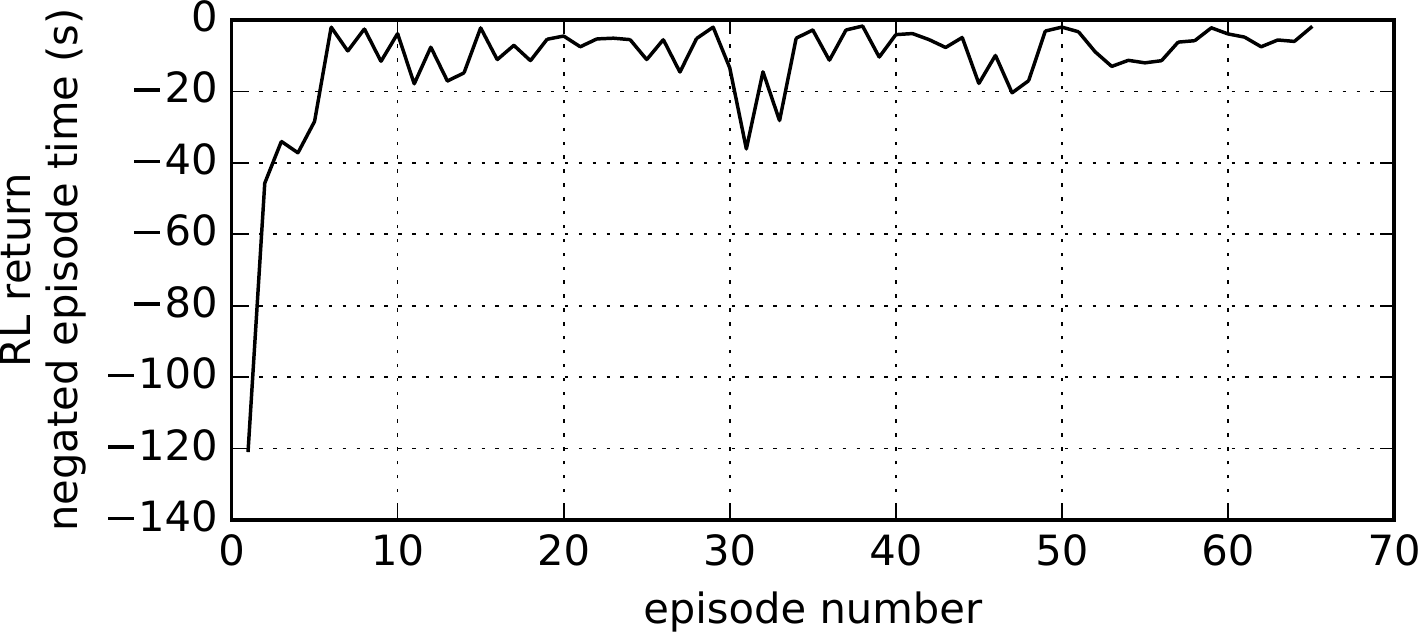}
  \caption{Learning curve of the RL agent. x-axis is the episode number and y-axis shows the RL return translated to the total time of each episode. Less absolute value of return means less episode time and better performance.}
  \label{fig:return}
\end{figure}

\begin{figure}[!t]
  \centering
  \includegraphics[width=0.75 \mycolumnwidth]{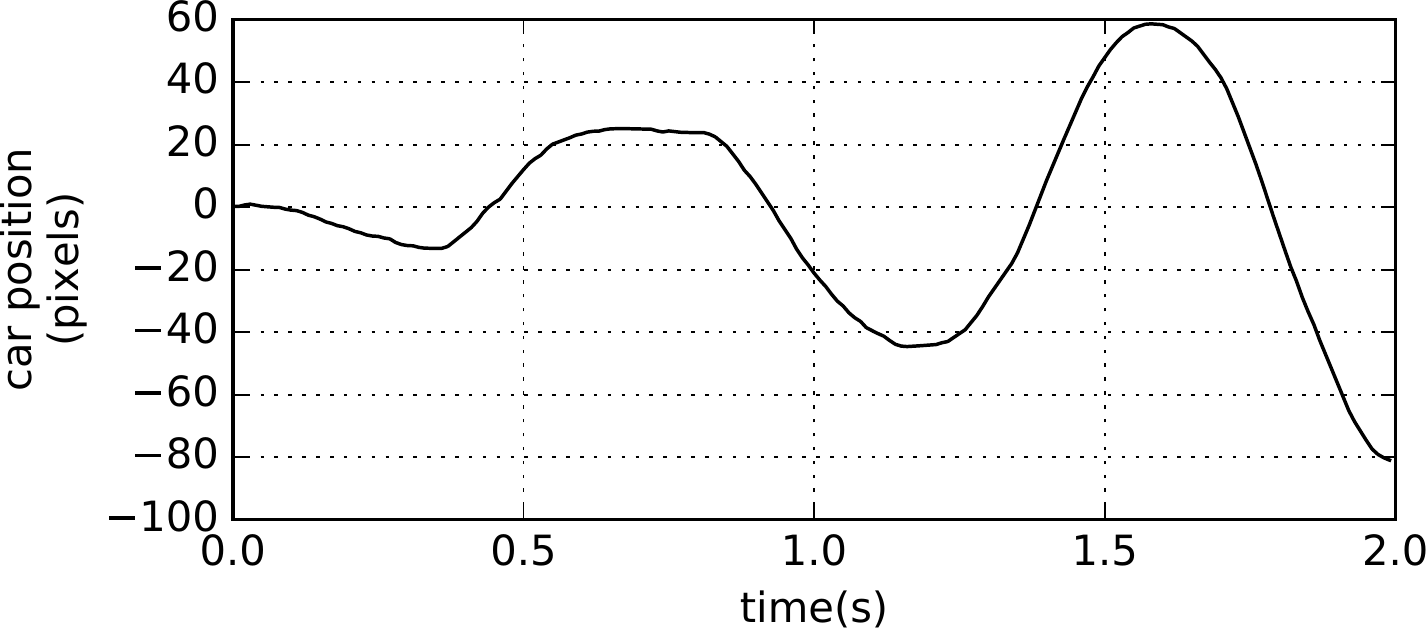}
  \caption{Car position vs. time plot obtained by RL algorithm after 37 episodes which its best performance in the experiment.}
  \label{fig:rlresult}
\end{figure}

Fig. \ref{fig:return} shows the learning curve of the AI agent where the accumulated return vs the episode number is shown. Fig. \ref{fig:rlresult} shows the learned policy at episode 37 which is the best performance obtained using the AI algorithm. The results show that the RL algorithm is able to achieve the performance of hand-engineered reference solution. The less number of swings made by the RL agent might be due to slight variations in the physical system and does not necessarily mean the superiority of the RL algorithm.

\vspace{-5pt}
\section {Conclusion} \label{sec:concl}
\vspace{-1pt}
In this paper, a novel physical environment benchmark is presented for AI algorithms. The environments can be implemented using low-cost parts and fabrication methods such as 3D printing. The proposed benchmarks enable researchers to easily replicate physical benchmarks to evaluate their AI algorithms and also share their implemented physical environments on the cloud with other users. Such collaborative benchmarking accelerates development of AI algorithms which can address challenges from real-world physical systems by engaging many researchers that can replicate the physical environments or access them on the cloud. We also presented an example implementation of the proposed physical environment framework. The results show the effectiveness of the proposed methods to develop a simple and low-cost physical benchmark.

Some possible future directions are adding more physical benchmarks, addressing the resource limitations of Raspberry PI for more computationally expensive algorithms and easy deployment of the whole framework on cloud solutions such as Amazon AWS.

\section{Acknowledgement}
This work was supported in part by the National Science Foundation under NSF grant number 1563652. 

\bibliographystyle{abbrv}
\bibliography{opeb}

\end{document}